\documentclass{article}

\usepackage{PRIMEarxiv}

\usepackage[utf8]{inputenc} 
\usepackage[T1]{fontenc}    
\usepackage{hyperref}       
\usepackage{url}            
\usepackage{booktabs}       
\usepackage{amsfonts}       
\usepackage{nicefrac}       
\usepackage{microtype}      
\usepackage{lipsum}
\usepackage{fancyhdr}       
\usepackage{graphicx}       
\usepackage{amsmath}
\usepackage{float}
\graphicspath{{media/}}     
\usepackage{subfig}
\usepackage{subcaption}
\usepackage{xcolor}
\usepackage{soul}
\usepackage{gensymb}
\title{A Machine Learning Approach to Generate Residual Stress Distributions using Sparse Characterization Data in Friction-Stir Processed Parts 
}

\author{
 Shadab Anwar Shaikh \\
  Reactor Materials Group \\
  Pacific Northwest National Laboratory \\
  Richland, WA\\
  \texttt{shadabanwar.shaikh@pnnl.gov} \\
   \And
   Kranthi Balusu \\
  Reactor Materials Group \\
  Pacific Northwest National Laboratory \\
  Richland, WA\\
  \texttt{kranthi.balusu@pnnl.gov} \\
   \And
  Ayoub Soulami \\
   Reactor Material Group \\
  Pacific Northwest National Laboratory \\
  Richland, WA\\
  \texttt{ayoub.soulami@pnnl.gov} \\
}

\begin{document}
\maketitle

\begin{abstract}
Residual stresses, which remain within a component after processing, can deteriorate performance. Accurately determining their full-field distributions is essential for optimizing the structural integrity and longevity. However, the experimental effort required for full-field characterization is impractical. Given these challenges, this work proposes a machine learning (ML) based Residual Stress Generator (RSG) to infer full-field stresses from limited measurements. An extensive dataset was initially constructed by performing numerous process simulations with a diverse parameter set. A ML model based on U-Net architecture was then trained to learn the underlying structure through systematic hyperparameter tuning. Then, the model’s ability to generate simulated stresses was evaluated, and it was ultimately tested on actual characterization data to validate its effectiveness. The model's prediction of simulated stresses shows that it achieved excellent predictive accuracy and exhibited a significant degree of generalization, indicating that it successfully learnt the latent structure of residual stress distribution. The RSG’s performance in predicting experimentally characterized data highlights the feasibility of the proposed approach in providing a comprehensive understanding of residual stress distributions from limited measurements, thereby significantly reducing experimental efforts.

\end{abstract}


\section{Introduction}

\quad Residual stresses are the "hidden" stresses that remain in a component even after the original cause of these stresses has been removed. They can arise at any stage of manufacturing or during the life of a metal component, with welding being a common source. Among the various welding methods, friction stir welding (FSW) and the associated friction stir processing (FSP) also generate residual stresses \cite{Balusu2023}\cite{baumann2014residual}, although typically to a lesser extent than traditional fusion welding \cite{moraes2019residual}. Despite their lower magnitudes, understanding residual stresses in FSW/FSP is crucial because they can still significantly impact a part's performance \cite{john2003residual}\cite{yadav2022combined}. Among these stresses, tensile stresses can lead to premature fracture under monotonic loading if present near flaws or other stress raisers \cite{chantikul1981micromechanics}; and they can enhance crack growth rates, which decreases fatigue life \cite{john2003residual}. Moreover, distortion caused by residual stresses can lead to misalignment issues, making it highly undesirable \cite{moraes2019residual}. The first step in studying residual stresses is determining their spatial distribution within the component rather than just focusing on their peak values. This is essential because, unlike in processes such as shot peening, residual stresses in FSW/FSP exhibit significant variation within the plane of the processed sheets \cite{woo2011neutron}. In FSW/FSP, the weld/process regions that undergo the most heating typically experience the highest tensile stresses, while other regions generally show either compressive stresses or negligible residual stresses. Due to the significant variation of residual stresses with in-plane location, the impact on the mechanical properties is also expected to vary considerably with location. Therefore, a comprehensive understanding of the full-field in-plane stress variation is crucial for studying residual stresses in FSW sheets. Note that although there is variation in residual stresses with depth \cite{woo2006angular} \cite{Balusu2024}, this study focuses on in-plane stresses because, through the thickness, the stress varies only in magnitude, not in sign (i.e., stresses are either entirely tensile or compressive).

\quad One method to determine residual stresses is through simulation, which has the added advantage of linking the process to the resulting stresses. Moreover, having a full-field simulated stress distribution facilitates their integration into mechanical performance studies, thereby completing the process-property-performance loop and achieving optimal performance. However, the issue with models is the inherent uncertainty in their predictions, which in turn could lead to issues with the reliability and credibility\cite{wang2020uncertainty}. There are several reasons for uncertainty. First, models inherently make assumptions about the physical phenomena occurring during the process. These assumptions are necessary for modeling a process like FSP, which involves multiple complex phenomena such as large deformations, large displacements, high strain rates, elevated temperatures, material property changes unique to this process, and large spatial gradients in all thermo-mechanical fields \cite{agiwal2022material}. Widely used heat source-based models for predicting residual stresses simplify the process to a heat flux input. Despite this simplification, these models have been validated in numerous cases, demonstrating their utility. \cite{chao2003heat}\cite{schmidt2008thermal}\cite{song2003thermal}. An alternative approach that is more representative of the process is to use the Smoothed Particle Hydrodynamics (SPH) numerical framework \cite{Balusu2022} \cite{LI2024108962}. SPH allows for considering more detailed friction stir processing (FSP) mechanics, such as material stirring, but simplifying assumptions regarding the phenomena at the boundary layer of the sample-tool interface remain unavoidable. Consequently, the uncertainty arising from the choice of the model and the simplifying assumptions encapsulated within it i.e. model form uncertainty persists across all modeling approaches. The second source of uncertainty is the parameter uncertainty that also exists in the two modeling approaches just discussed. In heat source models, parameters for the heat flux need to be determined, while in SPH models, parameters such as the heat transfer and friction coefficients at the sample-tool interface must be established. As a result, even if certain parameters are known to work for specific cases, there will always be uncertainty about the range or conditions under which these parameter values will yield accurate predictions. The co-authors have demonstrated one such case where the Thermal Pseudo-Mechanical (TPM) heat source model requires parameter modifications to remain effective beyond a certain range of friction stir tool shoulder radii.\cite{Balusu2023}.

\quad A more widely adopted approach is to measure the stresses using one of the various characterization techniques \cite{Withers2001b}. Each method, however, has its drawbacks. The contour method involves extensive destruction, suffers from low accuracy, and is time-consuming. Neutron diffraction requires a neutron source, making it feasible only in specialized labs. X-ray diffraction (XRD) offers higher spatial resolution. Still, it has a low penetration depth, necessitates careful sample and surface preparation, and is time-consuming for full-field measurements, especially when new material microstructures are encountered \cite{10.31399/asm.hb.v10.a0001761}. Synchrotron XRD offers a higher depth of investigation but is limited to only a few labs worldwide, making it less accessible. Hole drilling, while simple and capable of through-thickness measurements, still has drawbacks \cite{npl2517}. One issue is the reliance on strain gauges, which require considerable effort to set up and are susceptible to errors. However, electronic speckle pattern interferometry (ESPI) can overcome this limitation \cite{schajer2005fullfield}. In the presence of significant in-plane stress variations, it becomes necessary to characterize stresses at a large number of locations to capture this variation accurately. However, drilling many holes for these measurements is prohibitively time-consuming. An ideal solution would be a method that can infer the entire stress distribution from a limited number of holes. In essence, a residual stress “generator” that can generate the whole stress distribution when “prompted” with a few selected observations is required.

\quad This need to infer a complete residual stress distribution from a few selected measurements has been recognized in literature, and a few solutions have been provided. Without explicitly considering the process that produces residual stresses, most of these solutions assume the presence of incompatible inelastic strains, i.e., eigenstrains, lead to residual stresses \cite{mura1987micromechanics}. Determining residual stresses from eigenstrains is relatively simple and involves only linear elasticity calculations. However, the difficult task is to infer the eigenstrain distribution, i.e., the inverse eigenstrain analysis. Some inverse analysis methods rely on assumptions of distributions whose parameters were found through calibration to the available experimental sparse stress data \cite{chukkan2019iterative}. More generally, eigenstrain can be assumed to be given by a truncated series of basis functions with the coefficients as the unknowns found through fitting \cite{korsunsky2006residual} \cite{jun2010evaluation}. Typically, this approach would have resulted in an ill-posed problem with a non-unique solution, as the number of unknowns exceeds the available sparse characterization data. However, this is not an issue when either the distribution is simple or many aspects of its distribution are well understood. For instance, the residual stresses from peening are 1D, with most stress variations occurring near the surface, simplifying the problem significantly. As a result, the reduced number of unknown coefficients makes inverse eigenstrain analysis methods effective in these simpler distributions. Even in traditional welding, FE-based inverse analysis \cite{jun2010evaluation} approach was feasible due to prior knowledge about the stress distributions. This knowledge facilitated the determination of a characteristic length of stress variation, which, when combined with a 2D distribution assumption, significantly reduced the number of coefficients that needed to be fitted. However, an effective inverse eigenstrain approach for complex residual stress distributions, such as those found in FSP, is still lacking.

\quad Unlike eigenstrain methods, leveraging the mechanics' information from process simulations could potentially make inferring residual stresses from sparse data easier. For this purpose, machine learning (ML) models are promising. ML models have been used as surrogates for mechanics-based process simulations and bridge the gaps between different models \cite{shaikh2024finite}\cite{shaikh2024probabilistic}. Yang et al. \cite{yang2020prediction} predicted composite microstructure stress-strain curves using convolutional neural networks, highlighting their ability to consider spatial information while bridging length scales. Other examples of ML models considering spatial information include Bhaduri et al. \cite{bhaduri2022stress} who used a deep learning approach for stress field prediction in fiber-reinforced composite materials, while Hoq et al. \cite{hoq2023data} employed data-driven methods for stress field predictions in random heterogeneous materials. These studies underscore the effectiveness of multiple ML models in capturing complex spatial relationships in solid mechanics. In the context of sparse data reconstruction, Zhang et al. \cite{zhang2023improved} improved deep learning methods for accurate flow field reconstruction from sparse data, and Callaham et al. \cite{callaham2019robust} demonstrated robust flow reconstruction via sparse representation. Similarly, Xu et al. \cite{xu2023practical} utilized physics-informed neural networks for flow field reconstruction from sparse or incomplete data, indicating the promise of ML models in predicting full field from the sparse data.

\quad Given these successes, we propose an ML based approach that “generates” the full residual stress distribution when prompted with sparse stress characterization data. The ML algorithm will be trained on residual stress distribution data from a large number of process simulations with diverse process and model parameters. The inputs were selected sparse residual stress data points, and the output was the entire residual stress field. In other words, the ML model was used to capture the “latent structure” of the residual stress distributions \cite{shelhamer2017fully}. The specific deep learning architecture utilized was the U-Net. The U-Net, originally developed for biomedical image segmentation \cite{ronneberger2015u}, has shown remarkable performance in various image reconstruction tasks and forms the basis for diffusion-based models \cite{rombach2022high}. Its fully convolutional nature and ability to capture fine-grained details make it suitable for reconstruction tasks.

\quad This paper is organized as follows: First, the details of the finite element model used for simulating the FSP of the A380 sample are discussed. This is followed by the procedure to characterize the residual stresses and the details of the machine learning (ML) architecture, including its mathematical foundations, training, and testing. Subsequently, the results of the residual stress generator's (RSG) performance on the test datasets and unseen datasets, along with summary statistics, are detailed. Finally, the effectiveness of the RSG in predicting the full field stress from sparse characterization data is examined.

\section{Methodology}

\subsection{Simulation of Residual Stress Formation}\label{sec:sim}

The thermo-pseudo mechanical (TPM) model was utilized to simulate the residual stresses arising from friction stir processing \cite{Schmidt2005a}\cite{Schmidt2008}. This modeling approach employs a fully coupled thermo-mechanical finite element method implemented in the Abaqus/Standard software. Here, the material stirring is not explicitly modeled; instead, a simplified approach that treats the entire process as a moving heat flux is adopted. The heat flux, $q$, on the top surface of the plate, directly underneath the moving tool’s location, is expressed as: \begin{equation} q(r,T) = \dfrac{\zeta \omega r \sigma_{yield}(T)}{\sqrt 3}; \qquad \mathrm{for}:0 \leq r \leq R_{shoulder}\label{eq:1} \end{equation} where $\sigma_{yield}(T)$ is the yield stress as a function of temperature $T$, $r$ is the distance from the center of the tool’s shoulder, $\omega$ is the tool’s angular velocity, and $\zeta$ is the calibration factor \cite{Balusu2023}. This heat flux is defined using the user-defined heat flux subroutine - UFLUX. The values used for the temperature-dependent yield stress are listed in the Appendix. Figure \ref{fig:fea_model} shows the finite element model, which assumes symmetry across the processing line by modeling only half the plate's width (with the symmetry plane normal to the y direction). The processing direction (x direction) aligns with the longitudinal direction, which is the longest dimension of the sheet. Two 8-node cuboid temperature-displacement (DC2D4) mesh elements \cite{DassaultSystemesSimulia2014a} covering the thickness of the plate was utilized. To reduce the computation time, a coarse mesh was used in the direction of thickness. In addition, the element of size 1.5 mm (on each sides) within the plane of the plate was used. However, it should be noted that this model is not capable of accurately simulating the significant variation of stresses through the thickness. This simplification is justified, as the focus of this study is on the in-plane stress distributions.

\begin{figure}[H]
    \centering
    \captionsetup{justification=centering}    
    \includegraphics[width = 0.9 \textwidth]{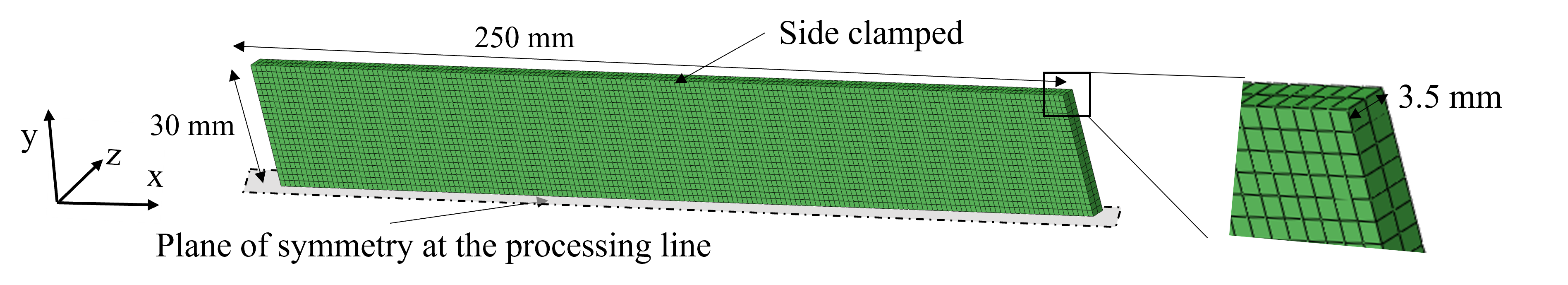}
    \caption{Finite element model for FSP of an A380 sample}
    \label{fig:fea_model}
\end{figure}

\quad Boundary conditions on the model include both displacement and thermal constraints. Initially, the displacement of the bottom surface was restricted in the thickness direction, to emulate the influence of a backing plate. Following this, the movement in the x-direction was constrained at the longitudinal ends of the plate. Finally, the transverse (in plane direction normal to the longitudinal direction) ends of the plate were fixed in all directions to imitate side clamps. These constraints remained through the processing step and were removed after the cool-down and in the unclamping step. Concerning the thermal boundary conditions, as most heat is known to dissipate through the backing plate, surface film conditions have been used to replicate this effect. A film coefficient value of 1000 W/m\textsuperscript{2} \degree{}C has been used.

\quad The simulations were run on a wide range of three parameter values: tool travel speed, tool rotation speed, and the calibration factor. These ranges were based on values typically employed and are displayed in the table below. The validation for one of the parameter sets used for training has already been done previously in one of the co-author's papers \cite{balusu2022the}. It is important to note that not all simulated data sets were used. Some simulations produced unrealistic process temperatures, either too low or too high for plausible FSP of aluminum alloys \cite{Threadgilll2009}. Therefore, simulations with process temperatures outside the approximate range of 400-500 \degree{}C were not considered, resulting in a total of 549 simulations. Of these, 400 simulations were used to create a dataset for training and testing the deep learning model. An additional 149 simulations were conducted to create a dataset for evaluating the generalization of the trained model. The parameter space differed between these two sets; the second set included mostly simulations using the two highest rotation speeds and calibration factors. Each simulation takes around 5-30 mins to run a single core of a personal computer. An important note is that, unlike the experimental sample, only one pass of FSP was simulated. This is because a one-pass simulation is expected to capture all essential features and patterns of the residual stress distribution needed to train RSG to learn the latent structure of stress distribution.

\begin{table}[H]
\centering
\caption{The simulation parameters used to generate stress data for training}
\begin{tabular}{ lcccc } 
 \toprule
 & Min. & Max. & Spacing \\ 
 \midrule
 Speed (m/min) &	0.01 &	1 &	0.1 \\ 
 
 RPM	& 300	& 1500 & 200 \\ 

 Calibration factor (${\zeta}$) & 0.1	& 0.5 & 0.05 \\
 \bottomrule
\end{tabular}
\end{table}

\subsection{Deep learning model}

\subsubsection{Theory}

From a mathematical standpoint, the learning task can be described by the equation:

\begin{equation}
\boldsymbol{\sigma}_{pred} = f( \boldsymbol{\sigma}_{sp}; \boldsymbol{\Theta}),
\end{equation}

where the goal is to construct a deep learning model $f$ that learns the mapping between the input, i.e., sparse observations $\boldsymbol{\sigma}_{sp} \in \mathbb{R}^{\mathrm{m \times m}}$, and the full residual stress distribution $\boldsymbol{\sigma}_{ref} \in \mathbb{R}^{\mathrm{m \times m}}$, where $m$ is the size of image-type data, by finding optimal hyper-parameters $\boldsymbol{\Theta}$. The model $f$ is said to learn when the prediction $\boldsymbol{\sigma}_{pred} \approx \boldsymbol{\sigma}_{ref}$. To facilitate learning, a dataset is constructed by assembling a set of inputs and their corresponding output observations. The model is trained by optimizing its parameters, typically using techniques such as back-propagation and gradient descent, with the goal of minimizing a loss function that quantifies the disparity between the predicted stress distribution $\boldsymbol{\sigma}_{pred}$ and the reference stress distribution $\boldsymbol{\sigma}_{ref}$. As the model learns, it adjusts its internal representations to capture the underlying mapping between the sparse observations $\boldsymbol{\sigma}_{sp}$ and the full residual stress distribution $\boldsymbol{\sigma}_{ref}$, ultimately achieving a prediction $\boldsymbol{\sigma}_{pred}$ that closely approximates $\boldsymbol{\sigma}_{ref}$ when evaluated on new data.

\subsubsection{Model architecture}

The proposed RSG was based on U-Net architecture. A U-Net initially developed for image segmentation purpose has proven its efficacy in capturing latent representations across various problems types \cite{cao2020denseunet}\cite{mei2021visual}. U-Net's standardized structure comprises of contracting layers followed by expanding layers, incorporating skip connections that propagate context information, ultimately enhancing output resolution. In this study, the encoder comprises of three repeating blocks, each featuring a 2 x 2 max-pooling operation for down sampling. Within each block, two consecutive 3 × 3 2D convolutions are applied with ReLU (rectified linear unit) activation function, followed by batch normalization operation. Notably, the encoder takes an input of size 128 x 128 and lacks the upfront max-pooling layer in the first block.

\begin{figure}[H]
    \centering
    \includegraphics[width = \textwidth]{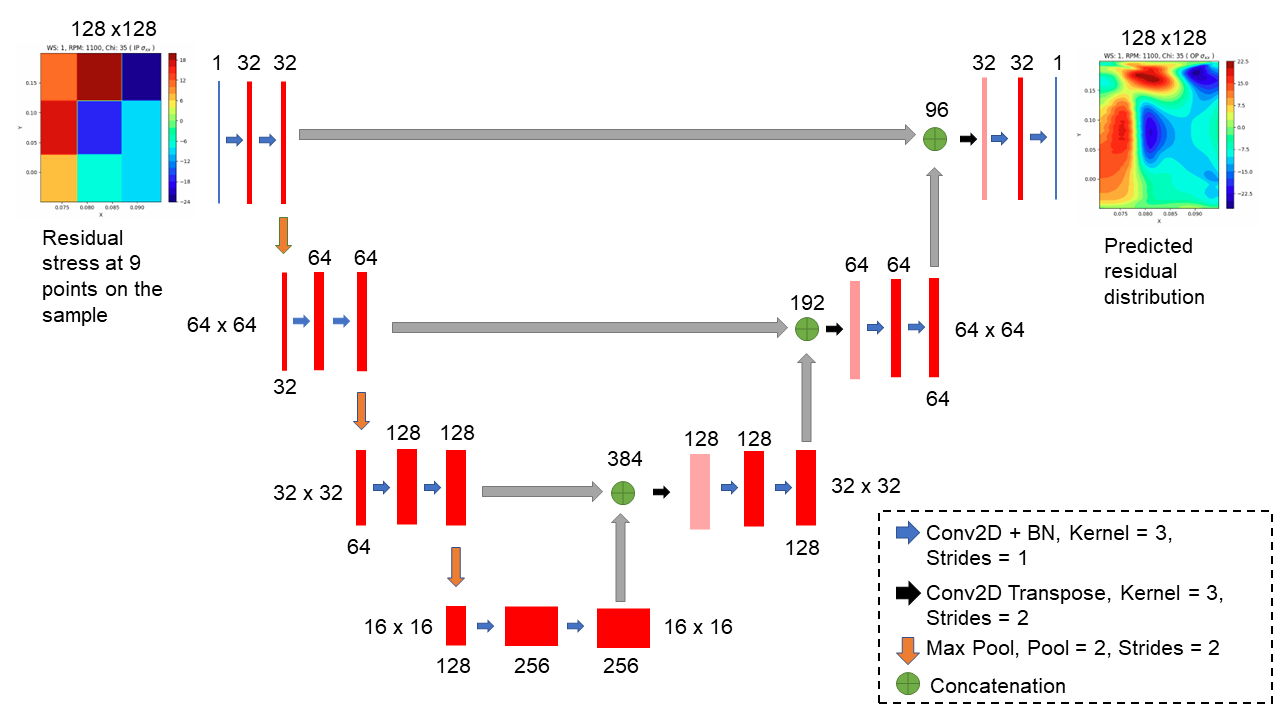}
    \caption{Schematic of RSG (U-Net) architecture used in this study}
    \label{fig:unet}
\end{figure}

The decoder segment consists of three repeating blocks, and an additional block at the start. The repeating block incorporates a 3 x 3 2D transpose convolution operation which receives a concatenated input from the preceding decoder and encoder block. Followed by two consecutive 3 × 3 2D convolutions with batch normalization and a ReLU activation. Notably, the first decoder block solely features the two 2D convolution layer, batch-norm, ReLu activation and max pooling from previous encoder block. The decoder blocks culminate in a final 1 × 1 convolutional layer (i.e. the output with size 128 x 128) mapping the 32-channel decoder output to a single channel. The training process involves minimizing the loss function, specifically the mean squared error, calculated between the predicted and true stress with the goal of learning the latent structure of stress distribution.  

\subsubsection{Data pre-processing}\label{sec:preprocessing}

In this study, data pre-processing was conducted on the data extracted (as comma separated values) from simulations to prepare it for training the RSG. The data consisted of coordinates of mesh centroids and corresponding value of thickness averaged longitudinal residual stress extracted at 2839 elements. The point data was interpolated onto a grid of size 128 x 128 using the grid interpolation functionality provided by SciPy \cite{2020SciPy-NMeth}, thus creating a 2D array of residual stress distribution from the simulation data. Subsequently, to simplify the learning task and improve convergence, the sparsity of the input was addressed by dividing the input into nine equivalent sub-regions. Additionally, these sub-regions were populated with the stress values at predefined sampling locations: (X, Y) = {(0.074 m, 0.010 m), (0.074 m, 0.080 m), (0.074 m, 0.170 m), (0.082 m, 0.010 m), (0.082 m, 0.080 m), (0.082 m, 0.170 m), (0.092 m, 0.010 m), (0.092 m, 0.080 m), (0.092 m, 0.170 m)}, all of which lie within each respective sub-region.

Later, the dataset was then curated by organizing these input-output pairs. i.e. input 2D array and a residual stress map. The dataset was then divided into training, testing, and validation sets. This division ensures that the deep learning model is trained on a substantial portion of the data, validated on a separate subset, and tested on a distinct dataset to assess generalization. Finally, the training, testing, and validation datasets were standardized using the Z-scoring technique, which involved scaling the inputs and outputs to have a zero mean and unit variance. This standardization promotes uniformity and aids convergence during the training phase.

\subsubsection{Evaluation metrics}\label{sec:eval_metrics}

In addition to the commonly used metrics such as RMSE (Root Mean Squared Error) and MAE (Mean Absolute Error), PSNR (Peak Signal-to-Noise Ratio) and SSI (Structural Similarity Index) were utilized in this study. PSNR and SSI were employed to objectively evaluate the quality of predictions, as has been similarly undertaken in several other studies \cite{strayer2022accelerating}\cite{kong2021deep}\cite{lee2023topology}.

PSNR provides a numerical assessment of how closely the processed signal resembles the original, with higher values indicating better fidelity, and is expressed in decibels (dB) \cite{hore2010image}. Similarly, SSI, a widely used metric in the fields of image processing and computer vision, quantitatively assesses the similarity between two images \cite{wang2003multiscale}. An SSI value close to one suggests a high similarity between the actual and reconstructed images. These metrics, coupled with RMSE and MAE, provide comprehensive insights into the quality of the reconstruction of the residual stress distribution.

\subsubsection{Training, testing and generalization}

The total dataset compromised of 400 input output pairs that was divided into training, testing and validation set following an 80\% - 10\% - 10\% split. The training subset consisted about 320 samples. Further, the testing and validation set had 40 samples each. The model was developed using TensorFlow library \cite{tensorflow2015-whitepaper} and was trained on NVIDIA Tesla V100 GPUs for 3000 epochs. Further, RSG consisted of about 2 million trainable parameters with structure illustrated in Figure \ref{fig:unet}. The optimal set of hyper-parameters such as learning rate, batch size was found to be 0.001, 16 respectively. 

The mean squared error (MSE) between the predicted residual stress and the training data was used as loss function to train the RSG. It is given by: \begin{equation}
\mathcal{L}=\frac{1}{n}\left\|\boldsymbol{\sigma}_{ref}-f\left(\boldsymbol{\sigma}_{s p} ; \boldsymbol{\Theta}\right)\right\|_2^2
\end{equation} where $n$ is the batch size used in training. The objective of the training process is to minimize the loss function to find the optimal model parameters. For each training epoch the loss is calculated for a given batch size and is back-propagated to the model. Further, the back-propagation algorithm updates the model parameters utilizing the gradients of the loss function. This process is continued until certain epochs such that the $\mathcal{L}\rightarrow 0$

Additionally, to be effective, a trained RSG should be able to predict residual stresses for new, unseen cases that are not part of the training and testing sets \cite{kochkov2021machine}. To evaluate its generalization ability, an additional unseen dataset consisting of 149 input-output pairs was generated. This dataset was built by performing process simulations on a unique parameter set different from the one used to generate the dataset consisting of 400 input-output pairs.

\subsection{Sample preparation and residual stress characterization}\label{sec:exp}
\begin{figure}[htbp]
    \centering
    \captionsetup{justification=centering}    
    \includegraphics[width = 0.9 \textwidth]{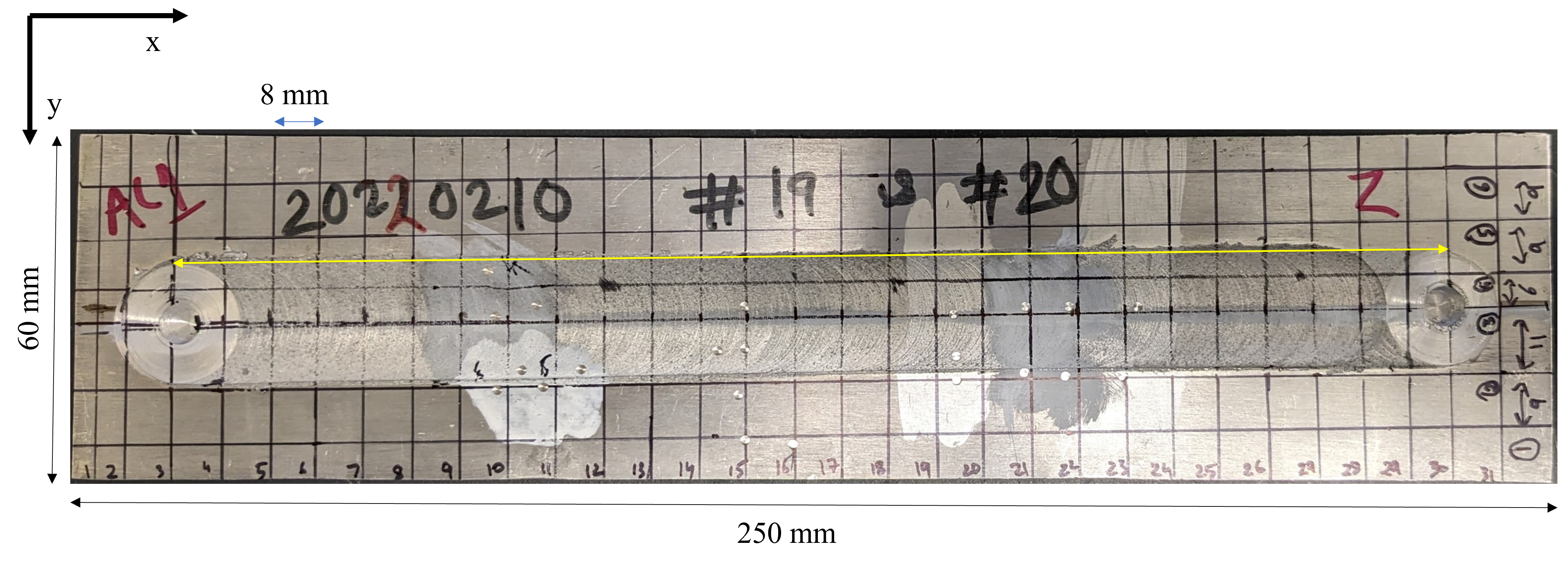}
    \caption{Top-view of the two-pass friction stir processed A380 sample. The vertices of the grid indicate the characterization locations}
    \label{fig:sample_snap}
\end{figure}

Two passes of friction stir processing were performed on a cast aluminum alloy 380 (A380.0) sample. A two-pass sample, as opposed to a one-pass simulation, was chosen to evaluate the RSG's generalization ability. The two passes were performed on the same straight line along the center of the sample but in longitudinal directions opposite each other. The sample was 3 mm thick and 250 mm long. The friction stir tool’s shoulder diameter was 20 mm. The tool rotates at 450 rotations per minute and moves at a 0.05 m/min speed.  More details of this process can be found in the paper from the coauthors’ collaborators \cite{Samanta2022c}. Fig. \ref{fig:sample_snap} shows the sample. On the same sample, two sets of parallel lines, each set being aligned with the two principal in-plane directions, were drawn. The spacing between the transverse lines is 8 mm, and in between the longitudinal lines, it is 6-11 mm. This variable longitudinal spacing ensures that, assuming the stresses are symmetrical on either side of the processing line, the spatial distribution of stresses is comprehensively captured. The minimum spacing of 6 mm—four times the hole diameter—ensures that the hole at an adjacent location does not influence the measurement. The intersection points for these line are where holes would be drilled for stress characterization. 

\begin{figure}[htbp]
    \centering
    \captionsetup{justification=centering}    
    \includegraphics[width = 0.5\textwidth]{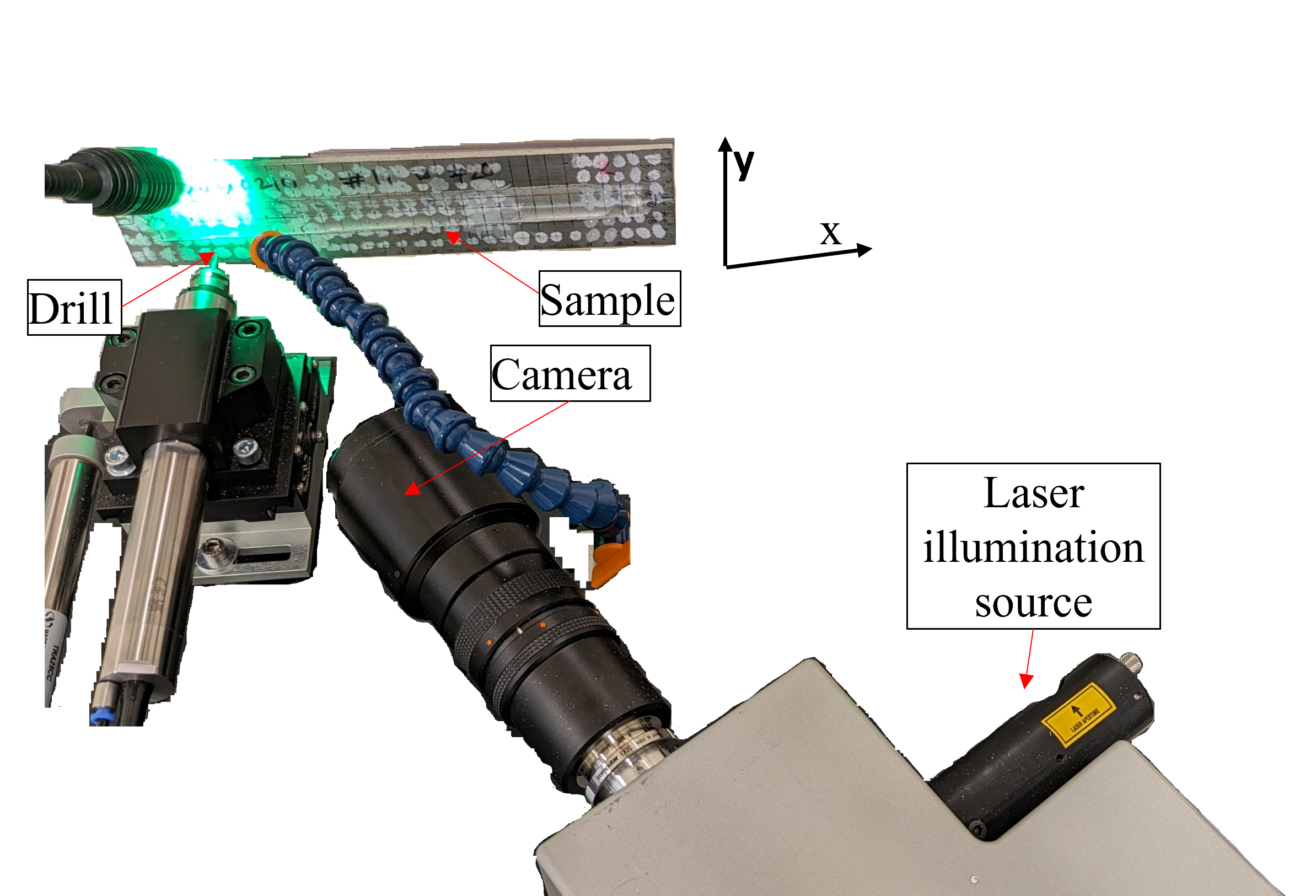}
    \caption{The hole-drilling electronic speckle pattern interferometry (ESPI) setup showing all the modules}
    \label{fig:ESPI_setup}
\end{figure}

Residual stress characterization was performed using hole-drilling electronic speckle pattern interferometry (ESPI) in the PRISM3 instrument \cite{Rickert2016}\cite{Steinzig2003a}. Fig \ref{fig:ESPI_setup} shows the components of the PRISM3 instrument. Hole drilling in a stressed component relieves some stresses, causing surface displacements near the hole. The associated surface changes are typically measured using strain gauges placed around the hole. In-plane stresses are then back-calculated using look-up tables populated with FEM simulations \cite{Schajer1981}. In contrast, this instrument uses ESPI to measure surface displacements. ESPI involves a mathematical operation using the laser-illuminated surfaces before and after hole drilling to determine relative surface displacement \cite{Schajer2005}, from which the stresses can be calculated.
By drilling at multiple locations, the 3D distribution of residual stresses in the sample can be inferred from through-thickness distributions at each location. A 1.5875 mm drill diameter was used, and since the allowable depth range is 0.1 – 0.5 times the diameter \cite{Steinzig2003c}, stresses at depths of 0.15 mm to 0.65 mm below the surface can be characterized. Each hole requires dedicated manual efforts, such as moving and reclamping the sample, readjusting the focus, recentering the drill, ensuring optimal image quality, and managing the automated incremental drilling and image analysis. As a result, it takes approximately 25 minutes per hole.

\section{Results}

\subsection{Simulation results}
\begin{figure}[htbp]
    \centering
    \captionsetup{justification=centering}    
    \includegraphics[width = 0.72\textwidth]{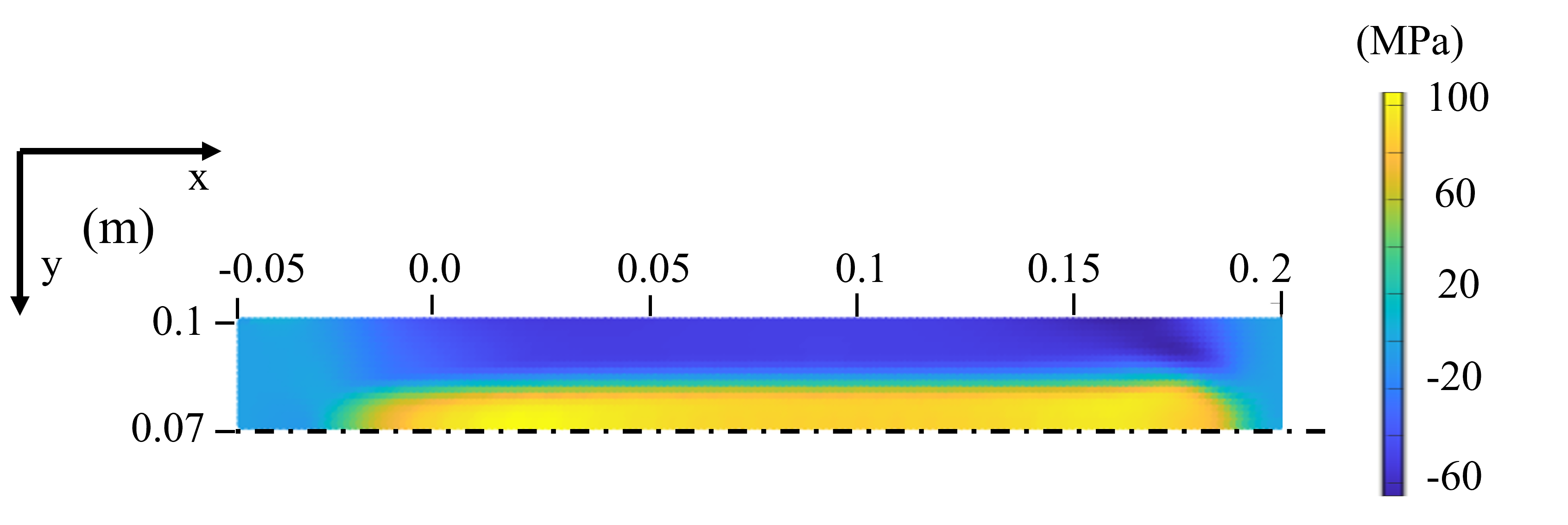}
    \caption{The half-model showing a simulated thickness-averaged longitudinal stress field used for training}
    \label{fig:sim_results}
\end{figure}

Considering the entire dataset of simulated results, a variety of stress distributions can be observed. These distributions share many key features, which will be described using the specific longitudinal stress distribution shown in Fig \ref{fig:sim_results}. This figure shows the thickness-averaged longitudinal stress field in the plane of the sample for a simulation of the process with a tool traverse speed of 410 mm/min, at 300 rpm and using a calibration factor of 0.15. Residual stresses show significant variation in the plane of the sheet, both along the processing line and perpendicular to it.  However, there is a region at the middle of the processing line where the stress variation in the processing direction is negligible, with variation occurring predominantly in the transverse direction. This transverse variation is often referred to as a plateau shape because the stresses are tensile in the stir zone, but they rapidly decrease to be compressive just beyond this region. The highest compressive values are observed at the plate edges. Moving beyond the middle region of the processing line, the residual stresses near the ends of the processing line differ significantly, exhibiting much higher tensile values than those observed elsewhere. Note that residual transverse stresses also exist; however, as observed elsewhere, they will be disregarded in this paper since they are much smaller \cite{Balusu2024}. Henceforth, longitudinal residual stresses will simply be called "stresses."

The rest of the simulation data set shares many qualitative features with the figure described. The primary difference lies in the stress values. The magnitude of the highest stresses ranges from about 20\% to 80\% of the base material's yield stress. There are also differences in the relative magnitude of the start and end zone stresses, but generally, these are higher than those in the center of the processing line. Additionally, there are minor variations in the shape of the stress distribution in the transverse direction. While the stresses near the stir zone remain tensile, the distribution is not always a plateau; the highest stresses may be located either at the center of the process zone or closer to the edge of the zone. 

\subsection{Prediction on test dataset}

The RSG received an input, that was prepared according to the methodology outlined in Section \ref{sec:preprocessing}. The model then predicts the stress distribution, which is quantitatively compared with the actual stress distribution using several evaluation metric discussed in the Section \ref{sec:eval_metrics}. The evaluation metrics such as PSNR and SSI were calculated using the sci-kit image \cite{scikit-image} Python library. 

The overall performance of the U-Net model on the test dataset (compromising of 40 samples) is summarized in Table \ref{tab:test_performance}. From the table, it can be inferred that the U-Net demonstrated excellent predictive accuracy, with RMSE and MAE values less than 4.3139 and 2.8385, respectively. Furthermore, the average values of SSI and PSNR were found to be 0.9810 and 40 dB, respectively, suggesting the model's efficacy in capturing spatial features, i.e., high structural similarity and less noisy predictions.

\begin{table}[htbp]
    \centering
    \caption{Summary statistics showcasing the performance of U-Net predictions on the test dataset}
    \begin{tabular}{lccc}
    \toprule
    Metrics / Values & Maximum & Average & Minimum \\
    \midrule
    RSME & 4.3139 & 0.9204 & 0.2279 \\
    MAE & 2.8385 & 0.6299 & 0.1743 \\
    PSNR & 52.8685 & 40.4053 & 27.7736 \\
    SSI & 0.9991 & 0.9810 & 0.9047 \\
    \bottomrule
    \end{tabular}
    \label{tab:test_performance}
\end{table}

For further analysis, the predictions of four samples, i.e., two best, and two worst performing, are shown in the Figure \ref{fig:result_test}. The differences between the actual and predicted stress distributions, $\Delta \sigma_{xx}$, were visually assessed through contour plot, providing an intuitive understanding of the model's performance in capturing the spatial features. As shown in Figure \ref{fig:result_test}(a) and (b), cases 1 and 2 showed an excellent model performance, with both having a lowest values of RMSE and MAE. Further, the PSNR values for these cases were above 50 dB, suggesting high-quality predictions i.e., lesser noise. Likewise, the SSI values were very close to 1, indicating a strong structural similarity to the actual stress distribution.
\begin{figure}[htbp]
    \centering
    \subfloat[]{\includegraphics[width=0.85\textwidth]{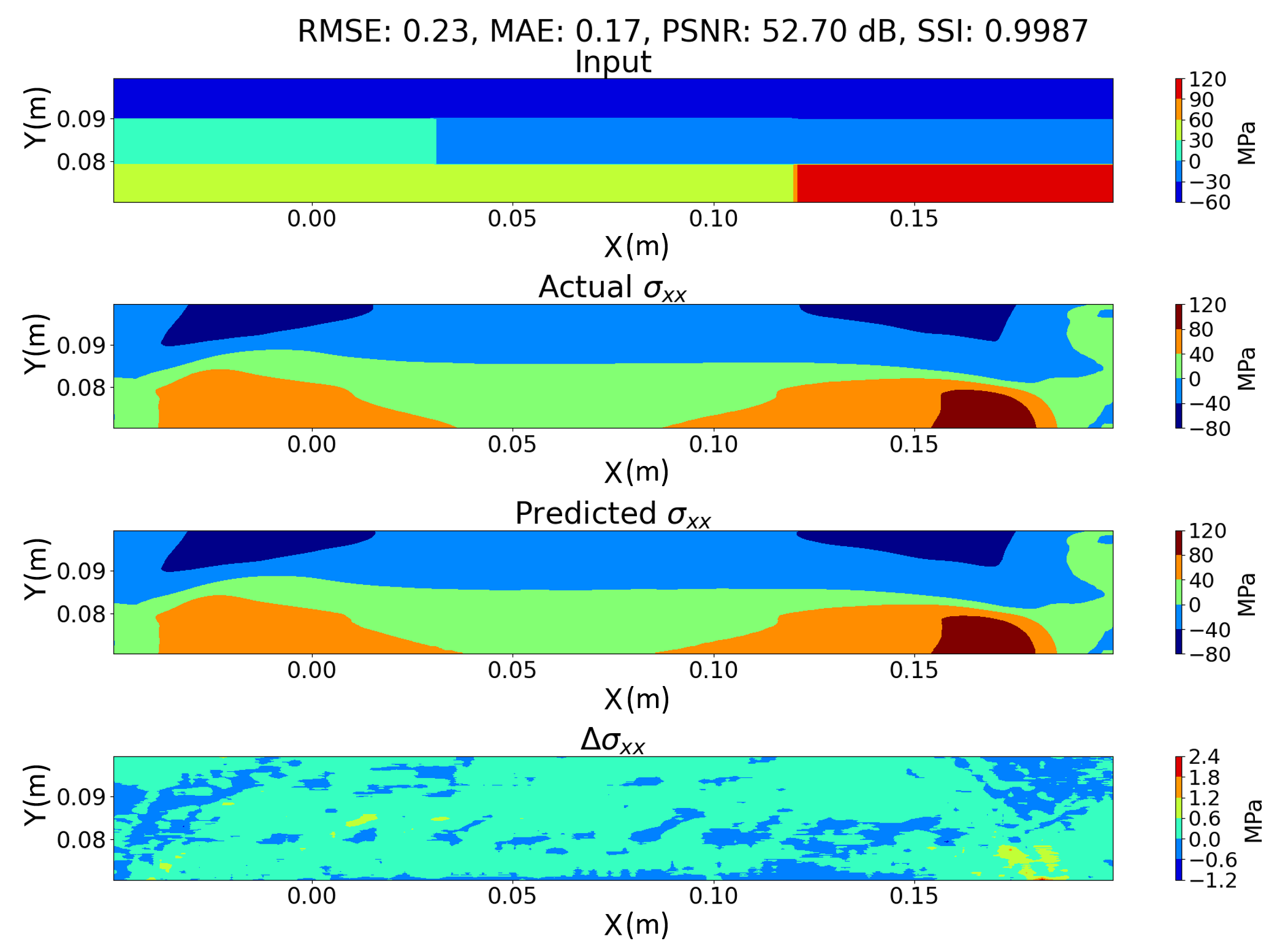}}\\
    \subfloat[]{\includegraphics[width=0.85\textwidth]{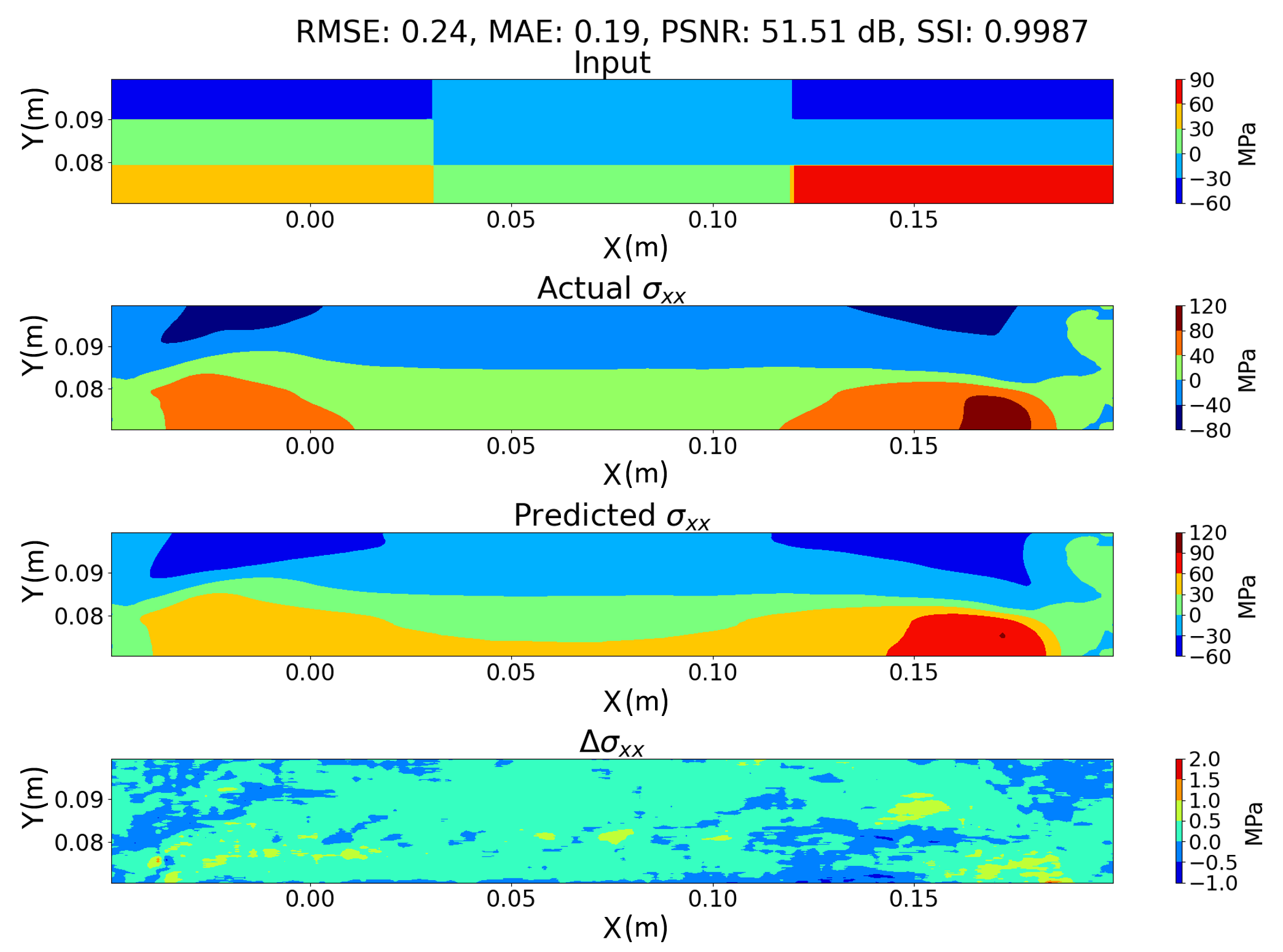}} 
    \caption{Predictions obtained using the trained RSG on the test dataset}
\end{figure}

\begin{figure}[htbp]
    \ContinuedFloat
    \centering
    \subfloat[]{\includegraphics[width=0.85\textwidth]{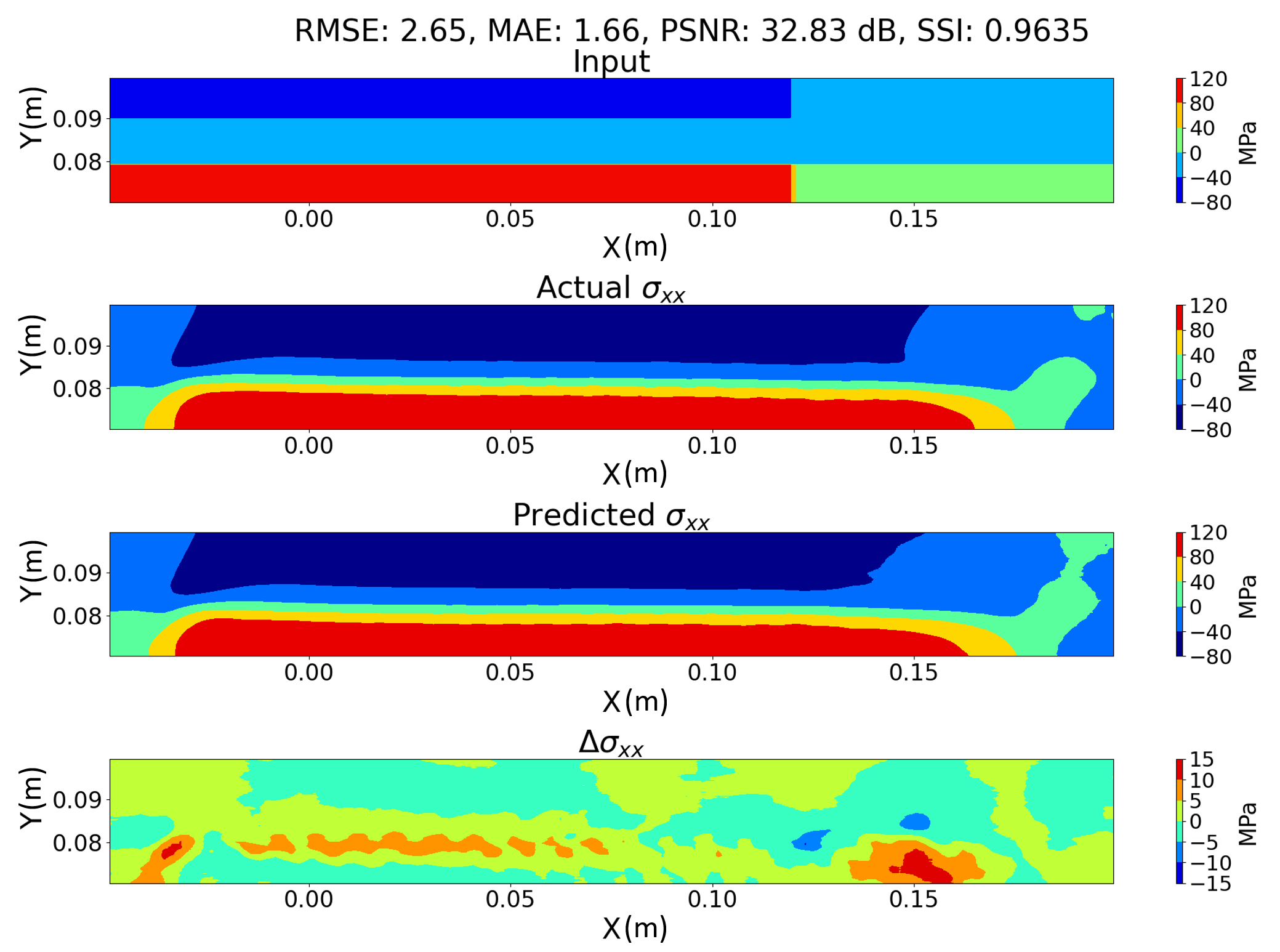}}\\
    \subfloat[]{\includegraphics[width=0.85\textwidth]{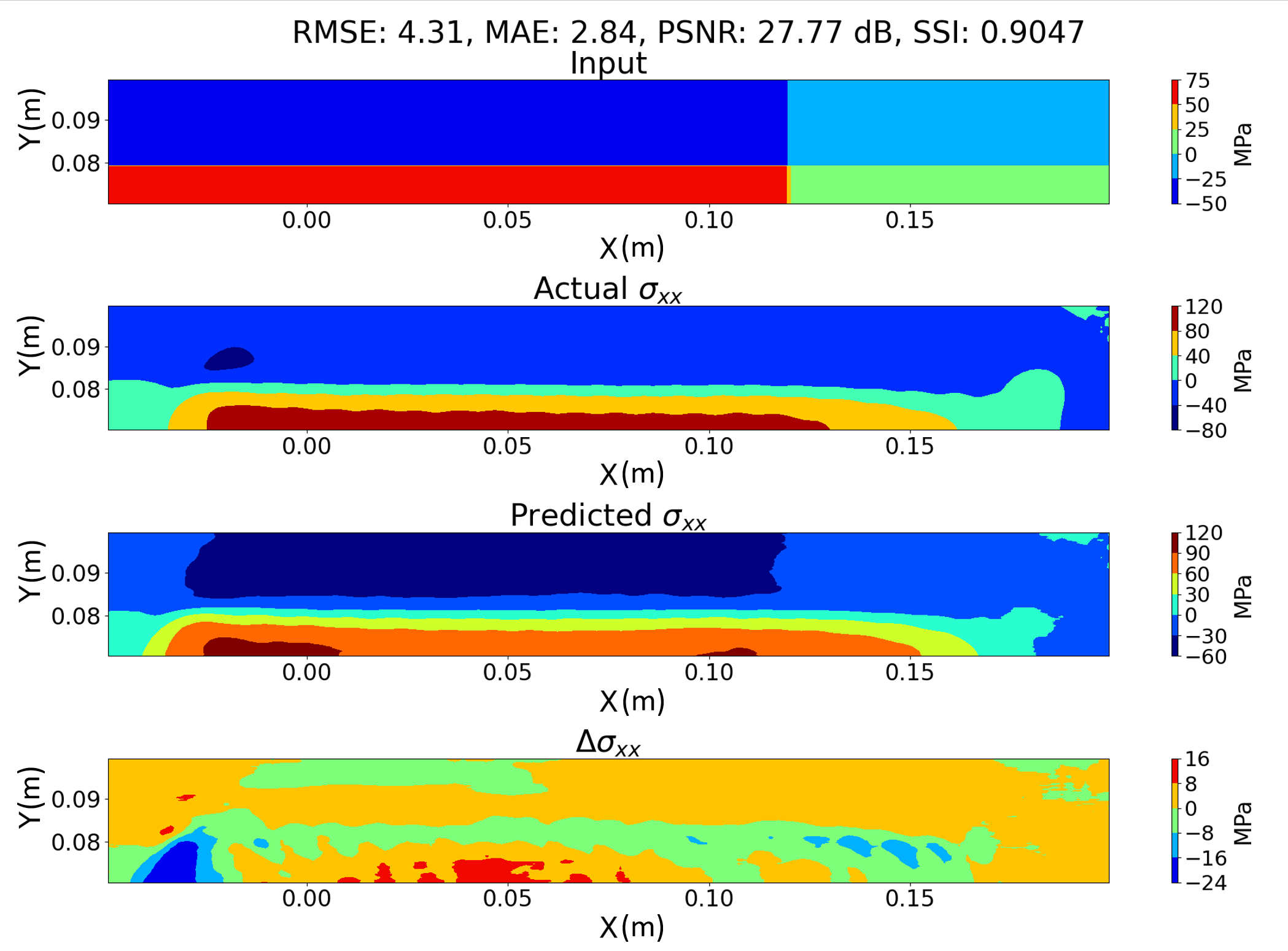}}
    \caption{Predictions obtained using the trained RSG model on the test dataset}
    \label{fig:result_test}
\end{figure}

On the other hand, cases 3, and  4 exhibited worst performance with highest values of RMSE and MAE. illustrated in Figure \ref{fig:result_test}(c), indicating a low prediction accuracy. This is also reflected in the lowest PSNR value and SSI, signifying that the predicted stress distribution had more noise and structural discrepancies compared to the actual distribution.

Visual inspection of the $\Delta\sigma_{xx}$ contour plots confirms the quantitative findings. Plots for cases 1 and 2 reveal minor discrepancies between the predicted and actual stress distributions, primarily at the center of the processing zone and sporadically in other areas of the map. Overall, these discrepancies remained minimal. In contrast, cases 3 and 4 display more pronounced discrepancies throughout the maps, especially in regions with high gradients, such as the ends and the edges of the processing zone.

\subsection{Test for generalization}

To asses generalization, the RSG was further evaluated on an unseen dataset, consisting of 149 input-output pairs generated by performing additional simulations. This dataset was pre-processed according to the methodology outlined in the Section \ref{sec:preprocessing} and was then provided to the pre-trained U-Net model.

The predictive performance results are summarized in Table \ref{tab:results_newdataset}. Additionally, predictions for four sample cases are shown in Figure \ref{fig:pred_newdata}.
\begin{table}[htbp]
    \centering
    \caption{Summary statistics showcasing the performance of RSG on unseen dataset}
    \begin{tabular}{lccc}
    \toprule
    Metrics / Values & Maximum & Average & Minimum \\
    \midrule
    RMSE & 16.9380 & 4.4264 & 2.7756 \\
    MAE & 13.9119 & 3.2440 & 2.1258 \\
    PSNR & 29.7148 & 25.0562 & 9.3580 \\
    SSI & 0.9446 & 0.8527 & 0.6321 \\
    \bottomrule
    \end{tabular}
    \label{tab:results_newdataset}
\end{table}

As outlined in the table, the RMSE and MAE for the entire dataset were found to be less than 16.9380 and 13.9119, respectively. The average SSI was 0.8527, with a minimum of 0.6321 and a maximum of 0.9446. Similarly, the PSNR values for the new dataset had averages of 25.0562, with minimum and maximum values of 9.3580 and 29.7148, respectively.

\begin{figure}[htbp]
    \centering
    \subfloat[]{\includegraphics[width=0.85\textwidth]{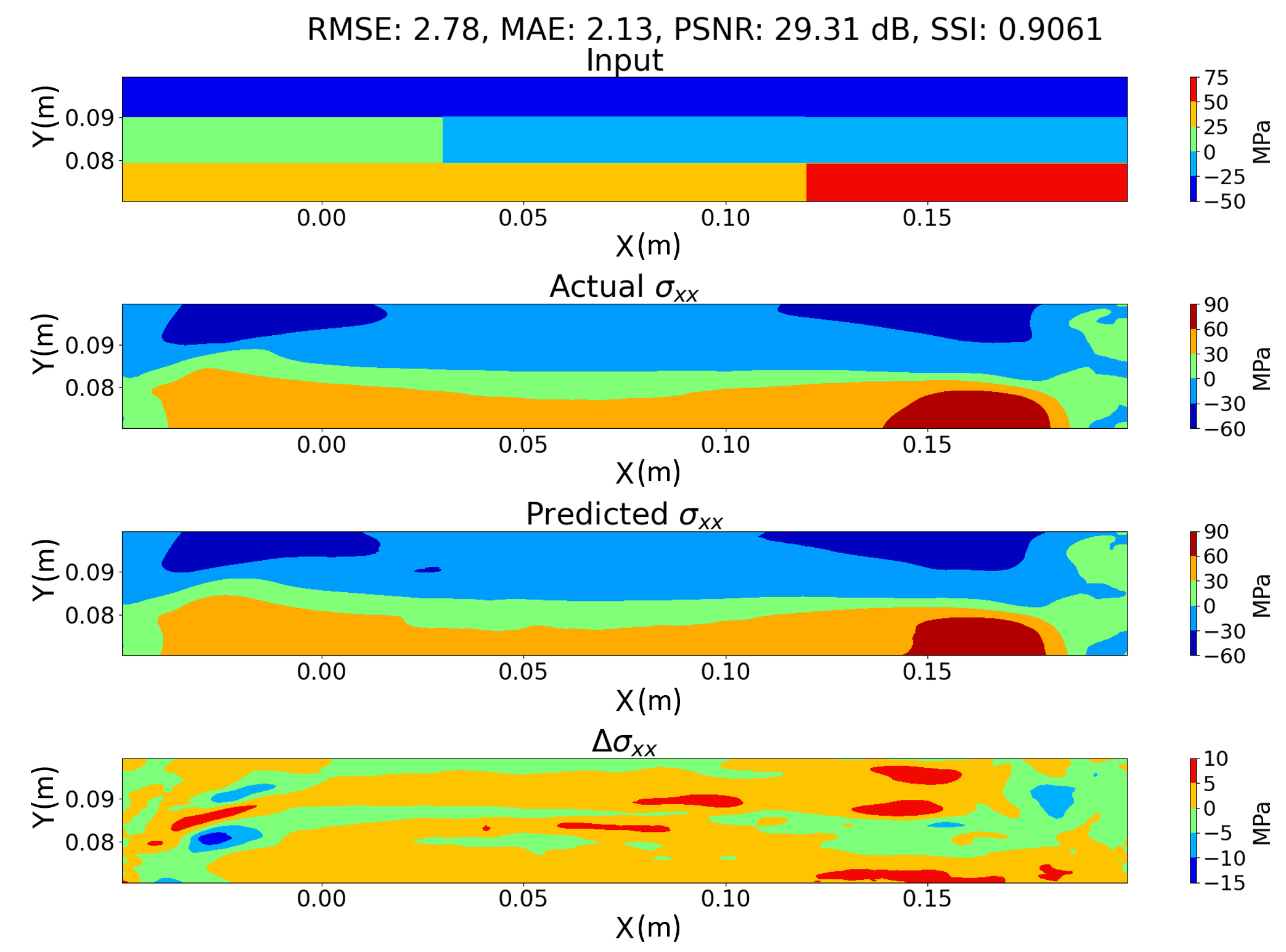}}\\
    \subfloat[]{\includegraphics[width=0.85\textwidth]{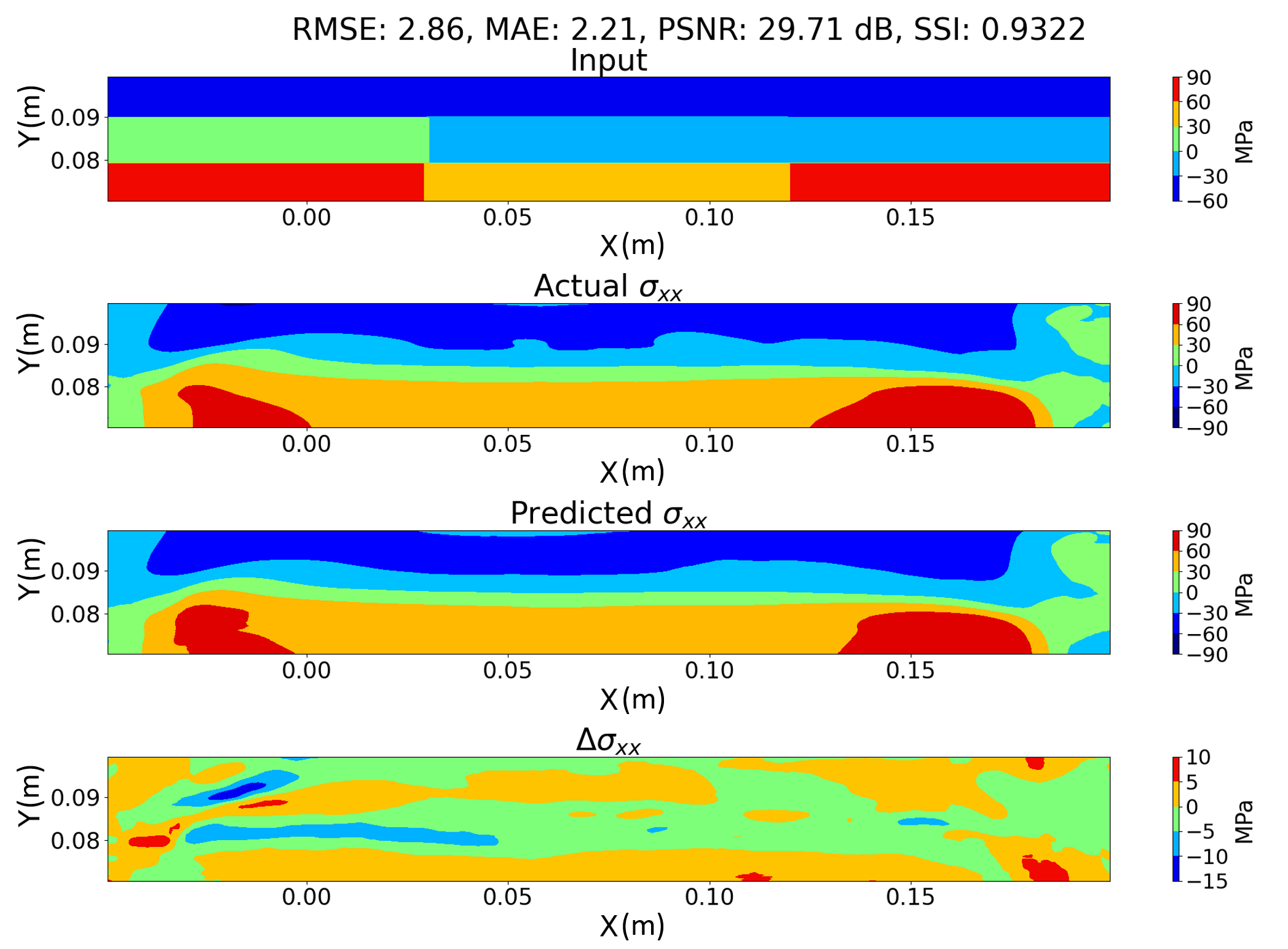}}
    \caption{Sample predictions obtained using the trained RSG on the unseen dataset}
\end{figure}

For sake of brevity, four random samples from the predictions on unseen dataset is shown in the Figure \ref{fig:pred_newdata}. From the visual inspection of $\Delta \sigma_{xx}$ contour plots, it is evident that the model's predictions closely match the actual stresses. This observation is further supported by the evaluation metrics.
\begin{figure}[htbp]
    \ContinuedFloat
    \centering
    \subfloat[]{\includegraphics[width=0.85\textwidth]{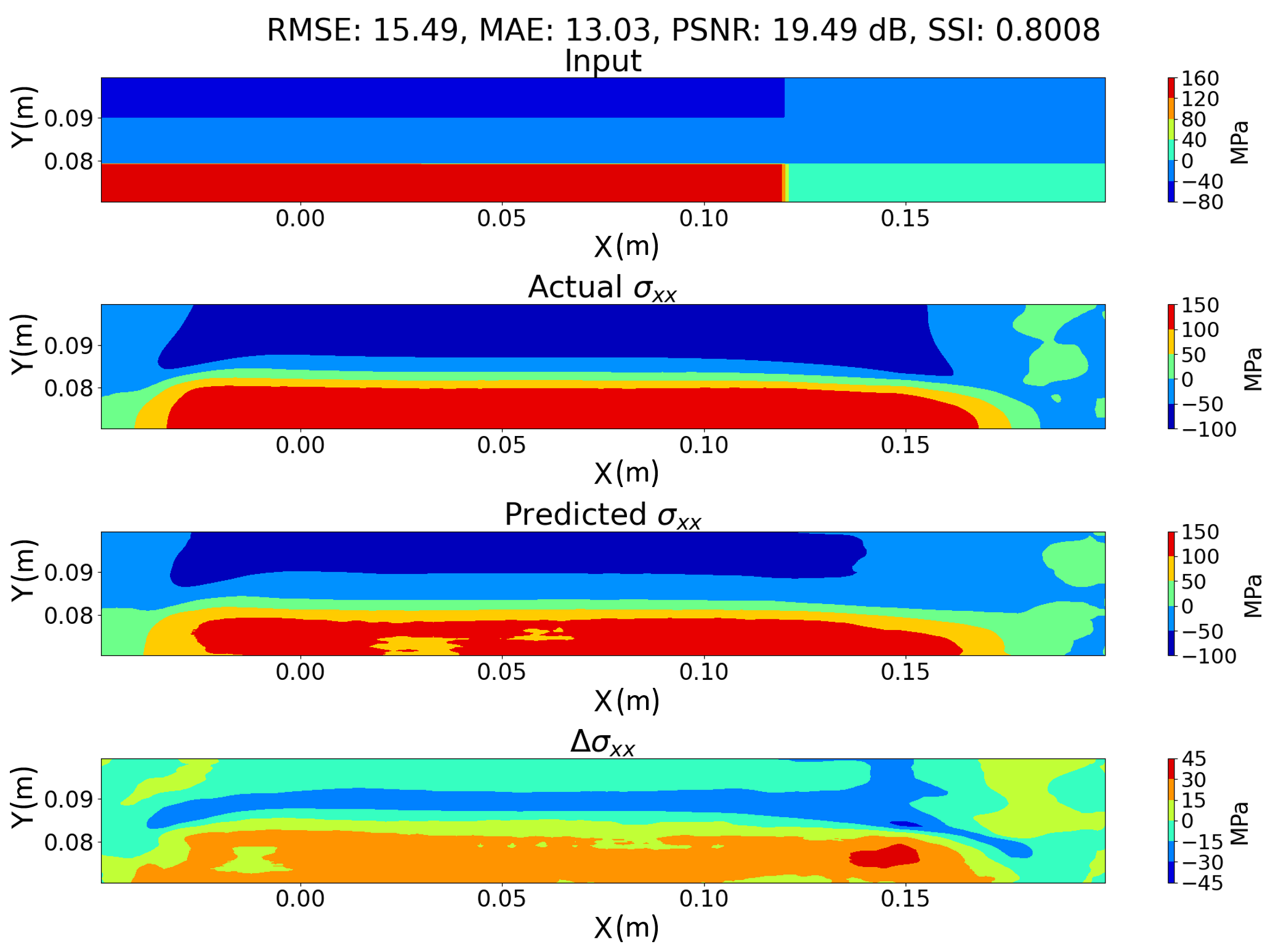}}\\
    \subfloat[]{\includegraphics[width=0.85\textwidth]{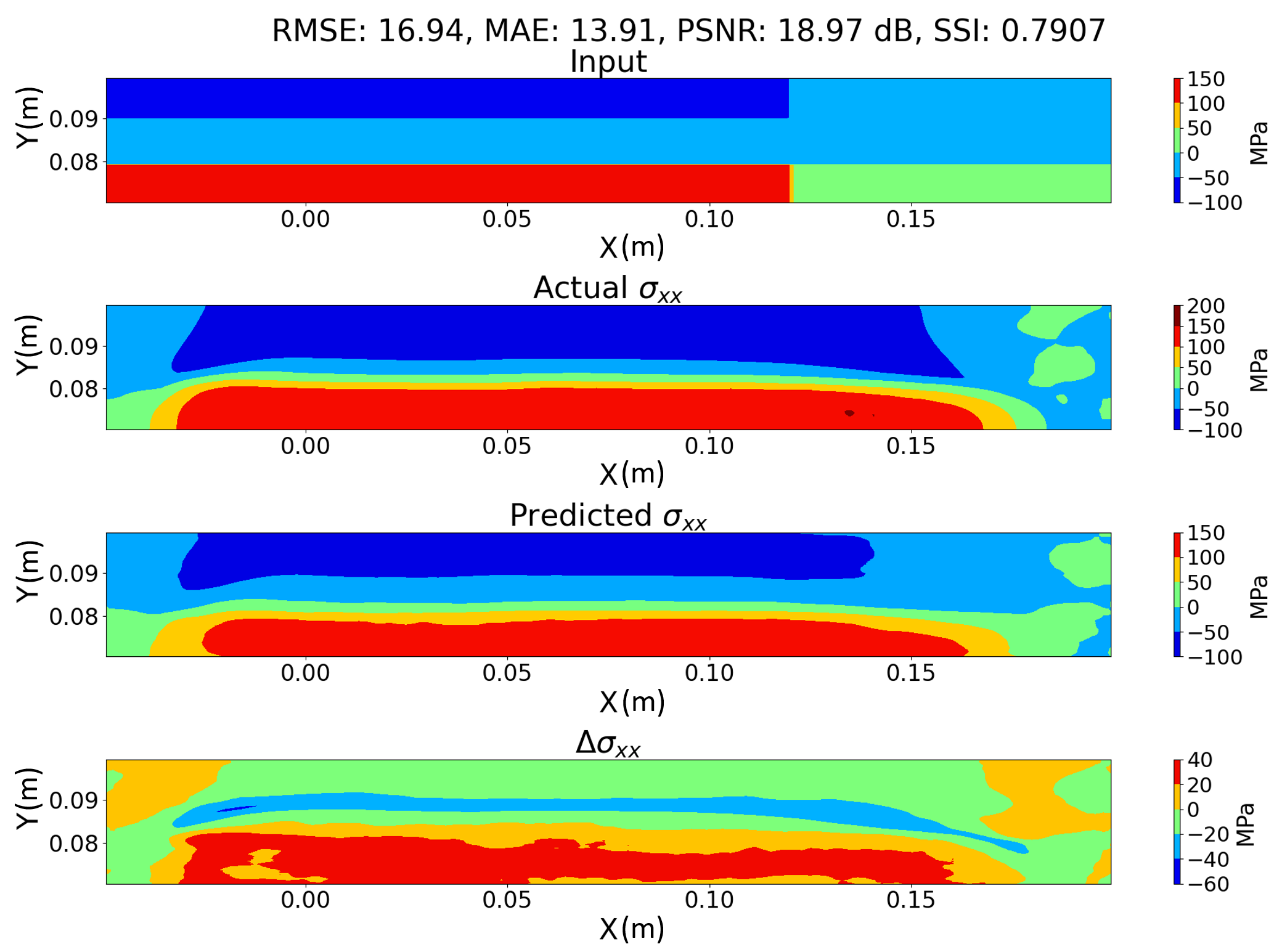}}
    \caption{Sample predictions obtained using the trained RSG on the unseen dataset}
    \label{fig:pred_newdata}
\end{figure}
Based on these observations, it can be concluded that the model demonstrated good predictive performance, accurately capturing the spatial distribution of residual stresses for most samples, as indicated by the average SSI. However, the predictions were found to be moderately noisy, as reflected by the low average PSNR values. Additionally, the model encountered challenges in accurately predicting the stresses, particularly in the processing zone and at several locations throughout the sample.

\subsection{Generation of stress distribution from characterization data}

\quad The capability of the RSG was further evaluated using characterization data from the FSP sheet. As shown in \ref{fig:Unet_vs_experimental_pred}(a), using the measured thickness average residual stresses at nine predefined locations as input, the RSG generated a full-field prediction of the stress distribution. The predicted stresses ranged from 80 MPa to -40 MPa, with the highest stresses near the start and end of the processing line (x = -0.025 m and x = 0.175 m) and at the edges of the processed zone (y = 0.08 m). Looking closer, stress variations exist in both the processing and transverse directions. Away from the ends of the processing line, the stress variation in the processing direction is small, except at the x = 0.04 m location. Consequently, the stresses at a specific longitudinal point far from the ends of the processing line, say at x = 0.075 m, can represent the stress variation in the transverse direction (y). Here, the stress varies significantly in the transverse direction: tensile stresses are observed within the processed zone (y < 0.08 m) and up to 5 m beyond it, while stresses closer to the edge of the plate are compressive. The highest stresses, around 46 MPa, were near the edge of the processed zone. Regarding the exception at x = 0.04 m, although the transverse distribution of stresses is similar—tensile in and near the processed zone and compressive elsewhere—the highest edge stresses at y = 0.08 m are notably lower.

\begin{figure}[H]
    \centering
    \subfloat[]{\includegraphics[width=0.9 \textwidth]{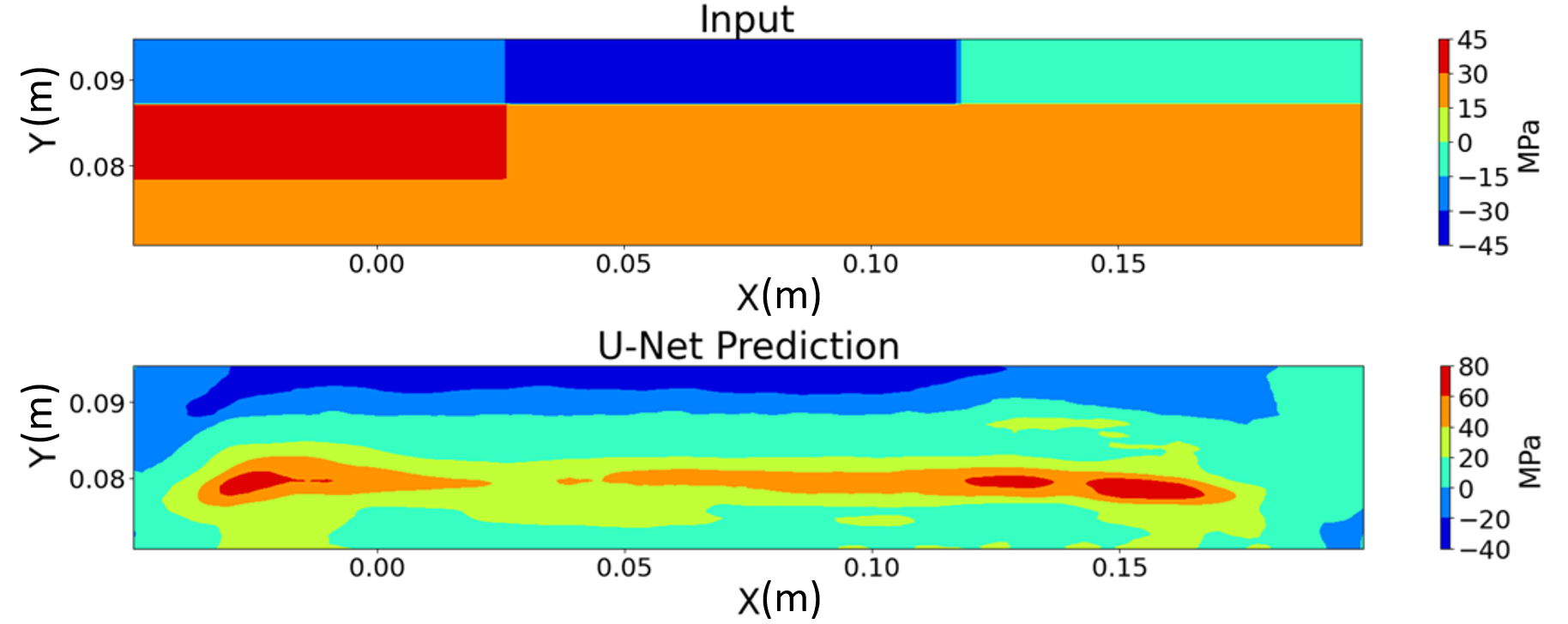}}
    \caption{Results form the trained RSG using the characterization data: (a) Displays the average residual stresses at nine predefined locations as an input to the model, alongside the predicted 2D stress distribution map.}
\end{figure}

\quad Figure \ref{fig:Unet_vs_experimental_pred} (b \& c) compares the model's predictions with the measured stresses, allowing for a quantitative assessment of its performance.  Although the original intent was to collect data from 186 locations, as indicated by the grid lines, only 120 of these locations were characterized due to experimental errors and omissions to conserve characterization efforts. The first step involved using a nearest-neighbor interpolation strategy to extract the predicted stress values at the characterization locations, as the RSG produced a 2D stress distribution map with 128 x 128 data points that do not coincide with the characterization locations on the sample. The overall model error, as indicated by the MAE and RMSE, was 10.59 MPa and 13.44 MPa, respectively. A more detailed analysis of the errors revealed that the model predicted residual stresses with an absolute error of 10 MPa or less at over 55\% of the locations, with approximately 70\% of these locations having errors of 6 MPa or less. The scatter plot in Figure \ref{fig:Unet_vs_experimental_pred} (b) provides a detailed analysis of the prediction errors by examining the relationship between the measured and predicted stress values, as well as identifying whether the errors are overpredictions or underpredictions. In this plot, the horizontal axis represents the measured stresses, the vertical axis represents the predicted stresses and the color of each point reflects the magnitude of the prediction error. The $R^2$ value, an indicator of correlation between the two data sets, is 0.6. The plot also reveals that most of the errors are overpredictions. Specifically, data points with errors exceeding 27 MPa are all instances of overprediction.

\quad The scatter plots in Figure \ref{fig:Unet_vs_experimental_pred}(b) are used to analyze the prediction errors in relation to their locations, providing insights into the accuracy of the stress distribution predictions at different locations. The plots display the measured residual stresses at all characterized locations, the corresponding RSG predictions, and the differences between them. The characterized data demonstrates that the qualitative features of the residual stress distribution, similar to those observed in the simulated stresses described in \ref{fig:Unet_vs_experimental_pred}(c), are consistent with the predictions made by the model. Examining the error values and their locations can quantify the accuracy. The regions with the largest errors can be grouped into three categories. The first region, located near the ends of the processing lines (at x = -0.025m and x = 0.175m), exhibits the most significant prediction errors. Except for one point, all locations with errors exceeding 27 MPa are within this region, and all of these are over-predictions. The second region includes two locations near x = 0.05m and close to the edge of the processing zone at y = 0.082m, which display the next highest errors. The simulation overpredicts these points by 25 MPa and 32 MPa. However, it should be noted that the characterized data at these locations shows very low-stress values, i.e. 3.5 MPa and 0.42 MPa, compared to nearby locations, where the values range from 15 MPa to 25 MPa.  The third region with notable errors is a horizontal line 5 mm outside the processing zone at y = 0.085m. Along this line, excluding areas near the ends of the processing region, the errors range from 11 MPa to 18 MPa. These errors are smaller than those in the previous regions and indicate under-prediction. Finally, with a few exceptions, all other locations outside of these three identified regions show errors of less than 8 - 9 MPa. The stress predictions at the center of the processed zone are the most accurate. Along this line, excluding areas near the ends of the processing region, the errors range from 0 MPa to -5 MPa, with one exception for the point at x = 0.071 m, whose error is 11 MPa. 

\begin{figure}[htbp]
    \centering
    \ContinuedFloat
    \subfloat[]{\includegraphics[width=0.4\textwidth]{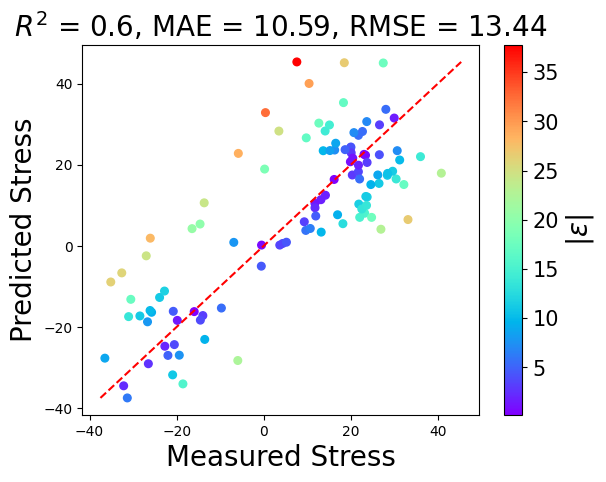}}
    \subfloat[]{\includegraphics[width=0.6\textwidth]{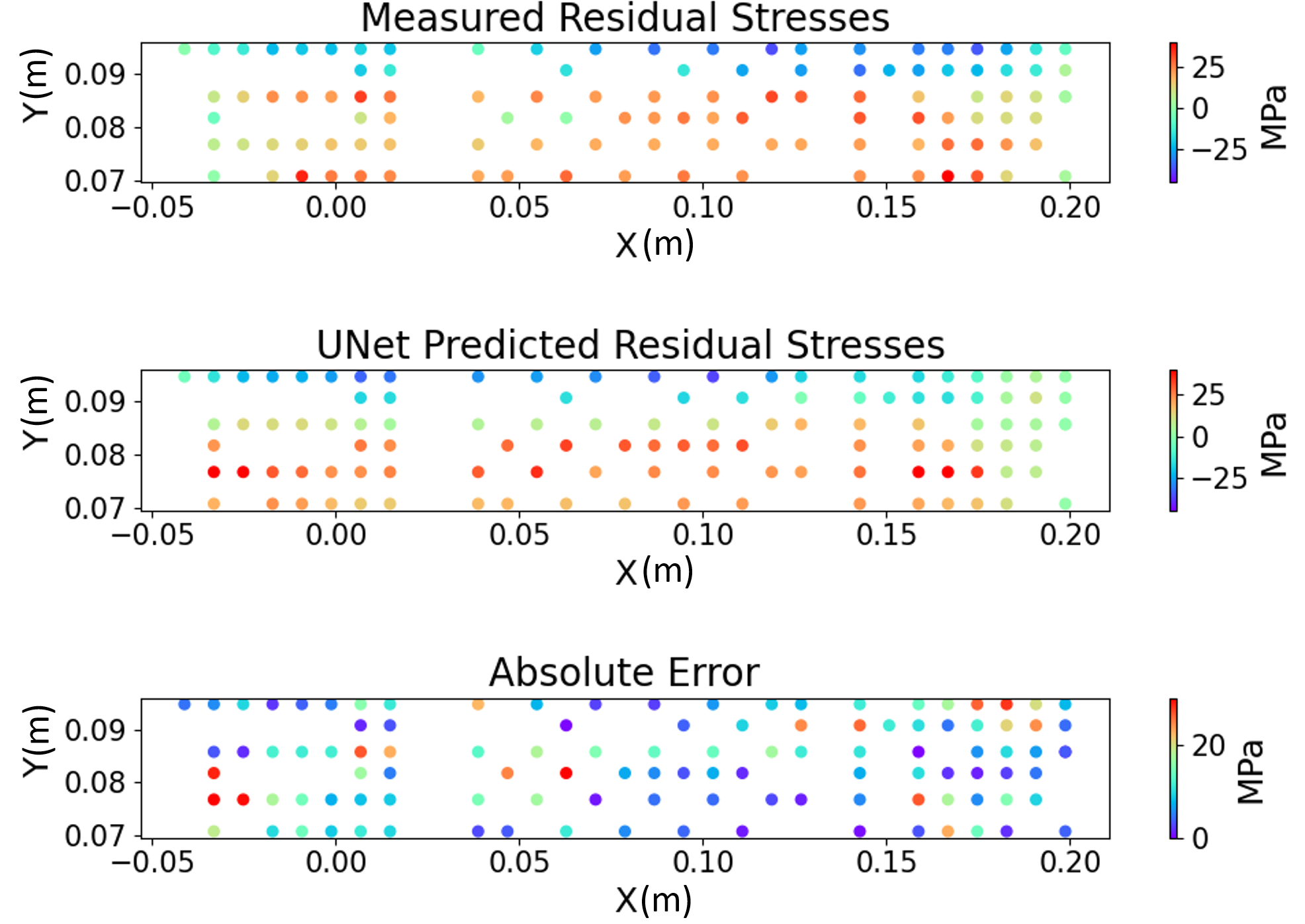}}
   \caption{Results form the trained RSG using the characterization data:(b) Illustrates the predicted versus measured stresses, with goodness of fit, and a prediction summary. (c) Offers a side-by-side comparison of the measured and predicted stresses across various sample locations.}
    \label{fig:Unet_vs_experimental_pred}
\end{figure}

\section{Discussion}

\quad The capabilities of the ML-based stress generator will be discussed in sequence, starting with its proficiency in generating simulated residual stress distributions. The error metrics for the predictions on the test dataset demonstrate the high accuracy of the residual stress generator. Specifically, the average RMSE and MAE values are less than 1 MPa, considering that the residual stresses are in the range of tens of MPa, which indicates excellent accuracy. Additionally, an average SSI value near 1 and a average PSNR of 40 dB confirm its efficacy in capturing spatial distribution. Further testing for generalization using a more extensive dataset shows only a slight degradation in performance. The mean errors increase to 3-4 MPa, and the SSI index remains close to 1 at 0.85. This result suggests that even on an unseen dataset, the accuracy of the generated simulated stress distributions remains high.

\quad The ultimate test for the residual stress generator is its ability to predict characterized stress distributions accurately. Qualitatively, the generated stress distributions successfully capture all known features associated with friction stir processing, as detailed elsewhere in the literature \cite{Balusu2024}. Specifically, the stress profiles exhibit the expected behavior: tensile near the process zone and highest near this zone’s edges and gradually transitioning to compressive further away. Additionally, the minor variations in stress along the process line, except at its ends, are accurately reflected. Quantitatively, the overall error metrics indicate moderately good prediction accuracy, with an $R^2$ value of 0.6 and an MAE value of 10.59 MPa. However, a closer examination reveals that these overall error metrics are influenced by a few outliers. Notably, 38.5\% of the locations have errors of less than 6 MPa, indicating that the stress predictions for a significant number of locations are accurate to within a few MPa.

\quad The outliers can be attributed to either measurement errors or the limitations of mechanics-based simulations. The discrepancies at the two locations categorized as the 2nd zone are likely due to measurement errors. These values are significantly lower than those of all characterized locations around them, suggesting a measurement anomaly caused by the unevenness of the area around these holes. Given that these holes are only 2 mm away from the edge of the process zone, there is a high likelihood of flash from material stirring exacerbating the surface unevenness, despite precautions taken to avoid this. Uneven surfaces pose a problem for the hole-drilling ESPI technique, which assumes a flat surface around the hole—up to three times its radius—for back-calculating stresses.

\quad The outliers in the other two zones can be attributed to the limitations of the training data sets and, therefore, the simulation models used. The outlier zone near the ends of the processing line can be traced back to the inaccuracies in modeling the tool's plunge and exit. The heat generation in these locations is not adequately captured by Equation \ref{eq:1}, which assumes the tool has fully plunged and represents steady-state conditions. This limitation cannot be addressed at present due to the lack of understanding of the residual stress formation and process conditions during the tool plunge and exit. In literature, studies on plunge and exit issues are rare \cite{Balusu2024a}, and residual stresses in these zones are only partially understood \cite{Balusu2024} \cite{Wang2022a}. Furthermore, as shown in Fig. \ref{fig:sample_snap}, the experimental sample undergoes two passes of FSP in opposite directions. This differs from the simulations, which assume only a single pass. Consequently, the experimental sample has no distinct plunge and exit locations. Additionally, the second pass of FSP does not align with the first pass; it was deliberately made shorter to avoid plunging at the exit hole of the first pass.

\quad The outlier in the 3rd zone can also be attributed to the limitations of the simulations. Here, the predicted stresses 5 mm outside the edge of the processed zone are 10 - 15 MPa lower than the experimental values, which were around 25 MPa. This discrepancy is not surprising. As shown in Figs.\ref{fig:sim_results}, \ref{fig:result_test}, and \ref{fig:pred_newdata}, all simulation results indicate low and even compressive stresses at this distance from the process zone. Consequently, the ML model trained on this simulation data understandably forecasts lower stresses at this location. The reasons behind this steeper transition from tensile to compressive stresses in simulations could be one of many. The clamping boundary conditions in the simulation may not accurately reflect the experimental conditions; this is indeed a known source of uncertainty \cite{Balusu2023}. Alternatively, the heat source distribution at the tool's shoulder edge may be inaccurate, or material hardening during the first pass of the FSP might play a role \cite{Balusu2024}. Ultimately, these simulation shortcomings are the underlying cause of the generator’s inaccuracies identified here.

\quad Identifying the reasons for the outliers serves two purposes. First, it provides opportunities to enhance the generator's overall accuracy. The issue of anomalous experimental data can be addressed by obtaining more reliable data or by “noisifying” the data. This later strategy has been successfully employed in multiple studies to make synthetic data resemble real-world data more closely when only a limited amount is available \cite{Weir2021}. Regarding, the error arising from the limitations of the simulations, it can be addressed in several ways. A straightforward way is augmenting the training data set to capture all sources of uncertainty more comprehensively, such as clamping conditions, which were not considered in this study. This includes considering the possibility of ineffective clamping, with varying extents of slipping zones in either of the two in-plane directions. A more involved approach is to improve the modeling to capture all physical phenomena during FSP better. One unaddressed phenomenon is how the process at the exit and plunge phases differs from the TPM model's assumptions. Future investigations could reveal how the process during these phases affects residual stresses, perhaps allowing for a simple modification to TPM model, such as adjusting the calibration factor or the heat flux area in \eqref{eq:1}, to be sufficient. Additionally, material hardening after the first FSP pass is another phenomenon that could be captured without abandoning the TPM model. As future studies provide more insights into microstructure evolution and its impact on material hardness, a more accurate model for the change in yield stress in \eqref{eq:1} can be incorporated. More generally, improvements to the simulation model need not be confined to simple modifications to the TPM model within the FEM framework. The generator's ability to use stresses at different locations as both input and output enables the seamless integration of all advances in mechanics-based simulations. Alternative numerical approaches with better representational accuracy, such as SPH, can also be employed to enrich the training data set.

\quad The second reason for identifying the causes of outliers is to systematically determine which components of the residual stress generator function most effectively. The development of the residual stress generator involved several aspects, including the selection of characterization locations to be used as input, the choice and training of the machine learning model, and the generation of the training data set, followed by rigorous performance evaluation. During the performance evaluation, analyses with both experimental data and simulation data confirmed that the first two aspects—the selection of characterization locations and the choice and training of the machine learning model—achieved excellent results.

\section{Conclusion}

\quad This study developed a machine learning-based residual stress generator (RSG) designed to determine full-field stress distributions from sparse characterization data. Notably, only simulation data was utilized for training due to the impracticality of obtaining large amounts of experimental data across various processing conditions. As the results showed, the generator accurately predicted stress distributions associated with FSP of Aluminum A380.0 sheets, achieving high accuracy with errors of about a few MPa for simulated distributions. Importantly, the model demonstrated its ability to generalize by providing good predictions ($R^2$ = 0.6) of an experimental stress distribution from a process markedly different from the training dataset. This performance underscores the model's capability to capture the latent structure of stress distributions. Consequently, by utilizing only 9 characterization locations to determine the full-field distribution instead of 120 locations—a more than 10x reduction—the RSG significantly reduces the experimental effort required to determine a full-field residual stress distribution.

\quad The ML-based approach introduced in this paper is intended to significantly enhance the capability to infer comprehensive information from limited observations. Without the RSG, deriving such insights requires substantial expertise in residual stress distributions, which is not widespread, especially with emerging processes FSP. This research underscores the transformative potential of machine learning in bridging the gap between mechanics-based simulations and experimental data to deliver a complete analysis of residual stresses. Furthermore, the implications extend beyond residual stresses, indicating potential applications for assessing the overall state of various components. Future research will focus on amplifying the accuracy and scope of the residual stress generator. Accuracy enhancement will involve expanding the training dataset to thoroughly integrate all uncertainty sources and refining the mechanics' models, generating the data to ensure they accurately capture relevant physical phenomena with high fidelity. Applicability will be increased by adapting the ML model to process 3D data and a diverse array of samples, allowing it to consider material properties, sample sizes, and geometries as variables. This development aims to make the model versatile and applicable to a broader range of scenarios beyond the current 2D stress distributions in flat sheets of a single alloy.

\section*{Acknowledgment and Funding}
The authors would like to acknowledge the role of Dr. Avik Samanta and Dr. Saumyadeep Jana in providing the friction stir processed aluminum A380.0 sample. The support provided by the U.S. DOE Vehicle Technologies Office and funding from the Light Metals Core Program (LMCP) is also acknowledged. Pacific Northwest National Laboratory is a multi-program national laboratory operated by Battelle for the DOE under Contract DE-AC05-76RL01830. 

\section*{Data Availability}
The data supporting the findings of this study will be made available upon request from the corresponding authors.

\section*{Conflict of Interest}
Author's declare no competing conflict of interest

\section*{Appendix}

\begin{table}[htbp]
    \centering
    \footnotesize
    \caption{Temperature-dependent yield stresses and other properties of aluminum alloy A380.0 \cite{Weir2021}}
    \begin{tabular}{ccccc}
    \toprule
    Temperature (\degree{}C) & Yield Stress (MPa) & Young’s modulus (GPa) & Specific heat capacity (J/Kg·K) & Density (Kg/m\textsuperscript{3})\\
    \midrule
    25	& 165 & 72 & 963 & 2770\\
    160 & 160 \\		
    200 & 146 & 63 & 1000 & 2760\\
    250 & 130 \\		
    300	& 77 & 53 & 1020 & 2730\\
    350	& 42\\			
    400 & 17 & 45 & 1100 & 2690\\
    \bottomrule
    \end{tabular}
    \label{tab:appendix}
\end{table}

Other material properties such as, poisons ratio, thermal conductivity, and thermal expansion coefficient were set to 0.33, 109 W/m·K, and 21.8 µm/m\degree C, respectively.

\bibliographystyle{unsrt}  
\bibliography{references}

\end{document}